\documentclass[runningheads]{llncs}

\usepackage{hyperref}
\usepackage{float}
\usepackage[T1]{fontenc}
\usepackage{graphicx}

\begin{document}

\title{An Experimental Study of Hybrid Retrieval, NLI and Open-source
LLMs for Scientific Claim Verification}

\author{Lakshmi Narayanan P\inst{1}\orcidID{0009-0006-7442-2144} \and
Narenkarthik K\inst{1}\orcidID{0009-0004-9499-2019} \and
Naren Kumar C\inst{1}\orcidID{0009-0009-2984-629X}}
\authorrunning{L. N. P. et al.}

\institute{Shiv Nadar University Chennai, Chennai, India\\
\email{lakshmi23110003@snuchennai.edu.in}}

\maketitle

\begin{abstract}
Claim verification is the task of deciding whether a source text
supports a claim, contradicts it, or lacks sufficient evidence.
Natural Language Inference (NLI) models can judge whether a single
sentence supports or contradicts a claim, but they struggle when the
evidence is spread across several sentences. Conversely, Large
Language Models (LLMs) can reason across sentences, but may draw
conclusions the text does not support. In this experimental study, we
test whether a conservative fusion rule, labeling a claim Support or
Contradict only when both models agree and defaulting to Not Enough
Evidence otherwise, combines their strengths. We construct a benchmark
of 2,035 claims from SciFact and CliniFact (858 Support, 152
Contradict, 1,025 Not Enough Evidence). We retrieve the top five most
relevant sentences per claim using a hybrid of semantic and lexical
(BM25) retrieval, which slightly outperforms either method alone.
These retrieved sentences are then provided to every verification
model. We then compare a DeBERTa-v3-base-mnli-fever-anli NLI baseline
against Llama 3.1 8B, Qwen3 8B, and Mistral 7B. Qwen3 achieves the
highest Macro-F1 (0.7937), followed by Llama 3.1 (0.7204) and Mistral
(0.6388), while the NLI baseline reaches only 0.3157. Contrary to our
hypothesis, fusion degrades performance, dropping Macro-F1 to 0.7392,
0.6730, and 0.5991, respectively. To understand this decline, we
analyze agreements, disagreements, and common error patterns between
the NLI and LLM verifiers. Our error analysis shows that NLI–LLM disagreements often result in correct LLM predictions being overturned.

\end{abstract}

\vspace{-5pt}
\section{Introduction}
\vspace{-5pt}

The rapid growth of scientific literature has made it increasingly difficult to manually verify the claims presented in research publications, particularly as individual claims may be supported or contradicted by evidence distributed across multiple sections of a paper. Scientific
claim verification is the process of evaluating whether a specific
statement or hypothesis made in scientific literature is supported or
contradicted by underlying evidence~\cite{wadden2020}. This
process requires a strict protocol, mainly in the retrieval of text and
in the reasoning done using this evidence.

Current verification systems rely on retrieval methods, natural language
inference (NLI) and LLMs. NLI models provide a constrained signal for
determining whether a claim is supported or contradicted by
evidence~\cite{kosprdic2024}, but may struggle when the required evidence
is distributed across multiple parts~\cite{yin2021docnli}. LLMs can
jointly reason over multiple pieces of retrieved context, but may produce
unsupported conclusions~\cite{ji2023survey}. These complementary
characteristics motivate combining their predictions through a
conservative fusion strategy, where a claim is assigned ``Support'' or
``Contradict'' only when both models agree. If not, the system abstains by
assigning ``Not Enough Evidence''. We hypothesize that this agreement-based
strategy can reduce mistaken verdicts by requiring independent agreement
between the two approaches.

This study makes three contributions. First, we establish a common evaluation framework to compare a pretrained DeBERTa NLI model with three open-source LLMs: Qwen3~\cite{yang2025qwen3}, Llama~\cite{grattafiori2024llama3}, and Mistral~\cite{jiang2023mistral7b}, using a shared hybrid retrieval pipeline on a benchmark of 2,035 claims labeled as Support'', Contradict'', or Not Enough Evidence''. Second, we evaluate three NLI evidence aggregation strategies and investigate a conservative agreement-based fusion strategy that returns Support'' or Contradict'' only when the NLI model and LLM agree, assigning ``Not Enough Evidence'' otherwise. Third, we conduct a comparative evaluation and detailed error analysis to assess the effectiveness and limitations of these approaches. Our results show that standalone LLMs outperform the NLI model, and that the agreement-based fusion does not improve Macro-F1 over standalone LLMs. The error analysis further examines the factors contributing to this outcome.

\vspace{-5pt}
\section{Related Work}
\vspace{-2pt}

\vspace{-5pt}
\subsection{Scientific Claim Verification Datasets}
\vspace{-5pt}
The task of scientific claim verification was introduced as determining
whether a scientific claim is supported or contradicted by the evidence
given in the literature. Wadden et al.~\cite{wadden2020} introduced the
SciFact dataset, consisting of literature from actual research, with
scientific claims paired with abstracts containing the evidence, along with
labels and the rationales supporting each verification decision. This work
established a benchmark for systems that retrieve relevant evidence and
reason over it, rather than relying on prior scientific knowledge.

Continuing this work, Wadden et al.~\cite{wadden2022} later introduced
SciFact-Open, which moves scientific claim verification into the open
domain. This extended the challenge from verifying claims against a defined
collection of evidence to finding relevant information in the broader
scientific literature.

While SciFact-Open focuses on broader scientific claims, CliniFact focuses
on claim verification in the clinical domain, where evidence for claims
derived from clinical research must be found in scientific
publications~\cite{zhang2025clinifact}.

\vspace{-5pt}
\subsection{Evidence Retrieval}
\vspace{-5pt}
Retrieving relevant evidence is the first step of a claim verification
pipeline. Lexical retrieval methods such as BM25~\cite{robertson2009probabilistic}
rank documents by their term overlap with the claim and remain a strong
baseline across domains. Semantic or dense retrieval
methods encode the claim and the documents into a shared vector
space, allowing relevant evidence to be found even when it is worded
differently from the claim~\cite{karpukhin2020dpr}.
However, dense retrievers trained on general-domain data can generalize
poorly to specialized domains such as scientific text. Hybrid retrieval fuses lexical and dense scores, for example through the combination of the two scores or reciprocal rank fusion, on the
belief that these two capture complementary notions of
relevance~\cite{bruch2023fusion}, and such combinations can improve over
standalone method alone. For scientific
claim verification specifically, Wang et al.~\cite{wang2025} used LLMs to
retrieve both corroborating and refuting evidence for a claim, highlighting
the importance of retrieving evidence on both sides of a claim.

\vspace{-5pt}
\subsection{NLI-based Verification}
\vspace{-5pt}
Natural Language Inference (NLI) determines the relation between a given
statement and a text given as evidence. It generally indicates whether the
evidence supports or contradicts the statement, or whether there is not
enough information to determine the relation. This allows NLI to closely
mirror the steps involved in scientific claim verification, where a claim
needs to be evaluated against evidence from research literature.

Košprdić et al.~\cite{kosprdic2024} used NLI for the task of scientific claim
verification. Their verification method includes a model fine-tuned on a
processed version of the SciFact dataset to determine whether a claim is
supported, contradicted, or lacks enough evidence. The authors fine-tune
multiple models, including RoBERTa Large, XLM-RoBERTa Large, DeBERTa, and
DeBERTa SQuAD, with the fine-tuned DeBERTa model achieving the highest weighted F1
score of 88\% on a held-out split of their processed SciFact data. Their verification engine is part of a larger scientific
question answering system, where documents retrieved from PubMed are used
as evidence for verification. However, NLI models are typically trained on
sentence-level pairs, and their performance degrades when evidence spans
multiple sentences or an entire document~\cite{yin2021docnli}.
They have also been shown to rely on shallow heuristics that do not
generalize beyond their training data.

\vspace{-5pt}
\subsection{LLM-based Claim Verification}
\vspace{-5pt}
The development of LLMs has opened a new avenue of approach to the task of
scientific claim verification, as these models can process a claim together
with the evidence text and generate the required verification decision.
Existing works have used LLMs for zero-shot and evidence-grounded
verification, including methods that take citation texts or retrieved text
as part of the input~\cite{alvarez2024}. These studies
use LLMs to perform claim verification reasoning without the need for a
task-specific classifier for every verification setting. However,
differences in prompting, model size, evidence selection and output
analysis make direct comparison between these LLMs
difficult~\cite{dmonte2026}. This motivates comparing multiple
open-source LLMs under the same retrieval and evaluation conditions.

\vspace{-5pt}
\section{Dataset Construction}
\vspace{-5pt}

\vspace{-5pt}
\subsection{SciFact}
\vspace{-5pt}

The first dataset we use is SciFact \cite{wadden2020}, a benchmark for
scientific claim verification. It contains scientific claims paired
with scientific abstracts as source text. Each claim in SciFact is
associated with a \emph{gold evidence} annotation, which identifies the
index of the sentence(s) in the corresponding abstract that provides
evidence for the claim. Claims that have gold evidence are labeled
``Support'' or ``Contradict'', and claims without it have no label.
\\

The dataset has training, development, and test splits. The test split
is excluded from our experiments because its labels are not publicly
available. We use the training split to build our benchmark and the
development split to select parameters.

We apply two filters. First, we remove claims whose gold evidence
spans multiple source documents, since our pipeline verifies claims
against a single source document. Second, we remove claims that have
no gold evidence and hence no label.\\

From the 809 training claims, 77 multi-document claims and 284 claims
without evidence and label are removed, leaving 448 claims: 296
``Support'' and 152 ``Contradict''.

The development split is used only to select the retrieval weights and
the NLI aggregation method, and it is not part of the final evaluation
benchmark. From its 300 claims, 24 multi-document claims and 110 claims
without evidence are removed, leaving 166 claims: 108 ``Support'' and
58 ``Contradict''.

\vspace{-5pt}
\subsection{CliniFact}
\vspace{-5pt}

The second dataset we use is CliniFact \cite{zhang2025clinifact}, a
dataset for evaluating clinical research claims paired with information
from scientific publications. It contains 1,970 claims. Each claim is
labeled ``Evidence'', ``Not Enough Information'' (NEI), or
``Inconclusive''.

All CliniFact claims come from a single source document, so we do not
need the single-document filter we used for SciFact.

We combine the training, validation, and test splits of CliniFact and
then map the ``Evidence'' label to ``Support'' and rename ``Not Enough
Information'' to ``Not Enough Evidence''. We exclude the 383
``Inconclusive'' claims, because our pipeline only outputs ``Support'',
``Contradict'', or ``Not Enough Evidence''. This leaves 1,587 claims:
562 ``Support'' and 1,025 ``Not Enough Evidence''.

CliniFact has no ``Contradict'' claims, so all 152 ``Contradict''
claims in our benchmark come from SciFact.

\vspace{-5pt}
\subsection{Constructed Benchmark}
\vspace{-5pt}

We combine the filtered SciFact and CliniFact claims into one benchmark
of 2,035 claims. Each claim has a single source document and one of
three labels: ``Support'', ``Contradict'', or ``Not Enough Evidence''.
The SciFact development claims used for tuning (166 claims) are not
included for evaluation.

Table~\ref{tab:benchmark} shows where each label comes from.
``Contradict'' claims come only from SciFact, and ``Not Enough
Evidence'' claims come only from CliniFact. ``Support'' claims come
from both datasets. The label distribution is imbalanced: ``Support''
makes up 42.2\% of the claims, ``Contradict'' 7.5\%, and ``Not Enough
Evidence'' 50.4\%.

\begin{table}[ht]
\caption{Label distribution of the constructed benchmark.}
\label{tab:benchmark}
\centering
\begin{tabular}{lrrr}
\hline
Label & SciFact & CliniFact & Total \\
\hline
Support & 296 & 562 & 858 \\
Contradict & 152 & 0 & 152 \\
Not Enough Evidence & 0 & 1,025 & 1,025 \\
\hline
Total & 448 & 1,587 & 2,035 \\
\hline
\end{tabular}
\end{table}

\vspace{-5pt}
\section{Methodology}
\vspace{-5pt}

\vspace{-5pt}
\subsection{System Overview}
\vspace{-5pt}

\begin{figure}[t]
    \centering
    \includegraphics[height=0.48\textheight]{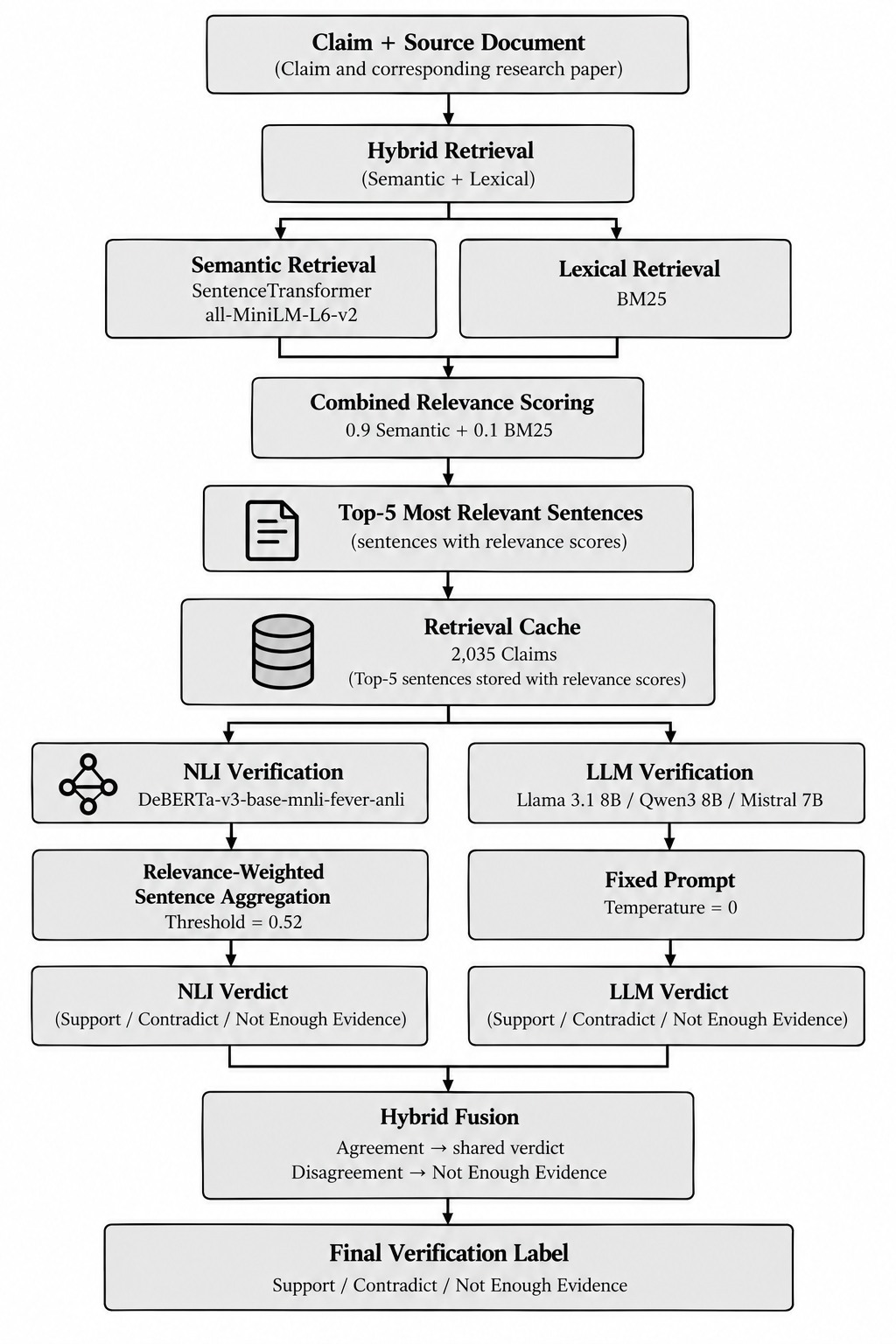}
    \caption{Overview of the claim verification framework.}
    \label{fig:pipeline}
\end{figure}

Given a claim and a source text, our experimental framework determines
whether the claim is ``Support'', ``Contradict'', or ``Not Enough
Evidence'' according to the source text. As shown in
Figure~\ref{fig:pipeline}, the framework is a multi-stage pipeline:
sentence retrieval, NLI-based verification, LLM-based verification, and
an optional fusion step that combines the NLI and LLM verdicts.

For each of the 2,035 claims, the top five most relevant sentences for
each claim--source pair are retrieved using a hybrid retrieval
configuration (Section~\ref{sec:hybrid-retrieval}). Subsequently, the
retrieved sentences and their relevance scores are stored in a
retrieval cache. All later steps read the top five sentences from this
cache, so every model operates on exactly the same retrieved data and
differences in performance come from the models, not from retrieval.

The NLI model (Sections~\ref{sec:nli_verification},
~\ref{sec:nli_aggregation}) and the three LLMs (Section~\ref{sec:llm_verify})
each use the same top five retrieved sentences to produce a verdict
independently. In the verdict fusion step (Section~\ref{sec:hybrid}),
the NLI and LLM verdicts are combined into a single final verdict.

\vspace{-5pt}
\subsection{Hybrid Retrieval}
\vspace{-5pt}
\label{sec:hybrid-retrieval}

Given a claim and its associated source text, the retrieval stage
identifies the sentences that are most relevant to the claim. We use a
hybrid retrieval approach that combines semantic similarity with
lexical matching to capture both semantic and exact-term
correspondence.

For semantic retrieval, we encode the claim and each source sentence
using the \texttt{all-MiniLM-L6-v2} Sentence Transformer model
\cite{reimers2019sentence,wang2020minilm}. The semantic similarity
between the claim and each sentence is computed using cosine
similarity. In parallel, BM25 \cite{robertson2009probabilistic} is
applied to the same set of source sentences to measure lexical
relevance based on term overlap.

Since the two retrieval methods produce scores on different scales, the
scores from both methods are independently normalized to the range
$[0,1]$. These normalized scores represent the relevance of each
sentence according to the respective retrieval method. We refer to the
normalized semantic similarity and normalized BM25 values as the
semantic relevance score and BM25 relevance score, respectively. The
two relevance scores are then combined using a weighted linear
combination:

\begin{equation}
\label{eq:relevance}
R_i = \alpha S_i + (1-\alpha)B_i,
\end{equation}

where $S_i$ is the semantic relevance score of sentence $i$, $B_i$ is
the BM25 relevance score of sentence $i$, and $\alpha$ controls the
contribution of semantic retrieval. The resulting $R_i$ is the final
relevance score assigned to sentence $i$ and is used to rank the source
sentences.

The parameter $\alpha$ is tuned on the SciFact development set. We
evaluate retrieval performance using Recall@$k$
($k \in \{1, 3, 5\}$) and Mean Reciprocal Rank (MRR). Recall@$k$
measures the proportion of claims for which at least one gold evidence
sentence is successfully retrieved within the top $k$ candidates. MRR
evaluates ranking quality by averaging the reciprocal rank ($1/rank$) of
the first retrieved gold evidence sentence across all claims in the SciFact development set.
Thus, higher MRR indicates that a relevant sentence is generally
positioned closer to the top of the ranking.

\begin{figure}[ht]
    \centering
    \includegraphics[height=4.5cm,keepaspectratio]{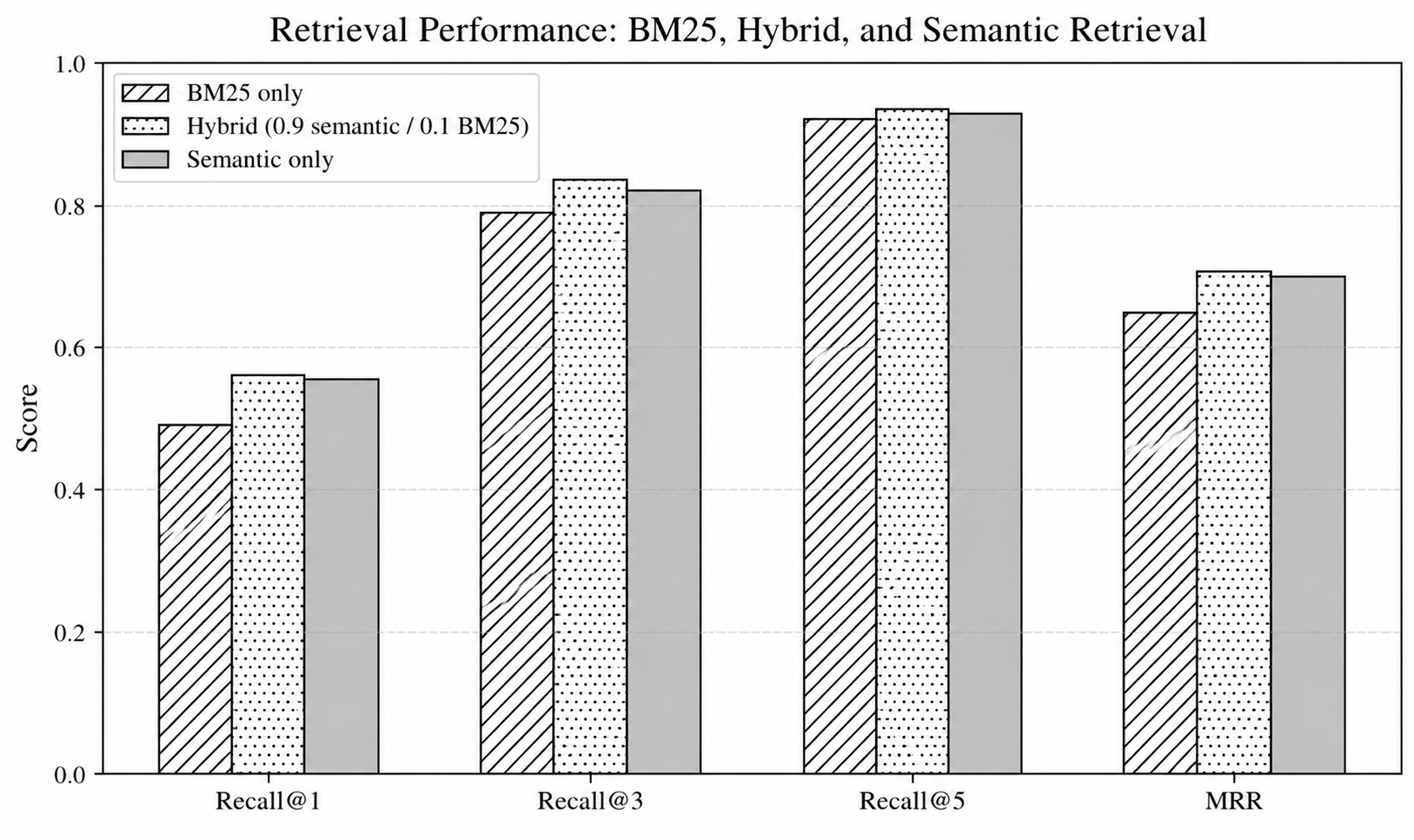}
    \caption{Retrieval performance comparison showing the hybrid model
    (0.9 semantic / 0.1 BM25) outperforming plain BM25 and plain
    semantic retrieval baselines.}
    \label{fig:retrieval-comparison}
\end{figure}

We test semantic weights ($\alpha$) from 0.0 to 1.0 in 0.1 increments,
with the BM25 weight set to $1-\alpha$. The $\alpha=0.9$ configuration
yields the highest MRR (0.7097) and mean recall (0.7811, calculated as
the average of Recall@1, @3, and @5) as shown in
Table~\ref{tab:retrieval_weight_selection}. At this optimal setting,
Recall@1, @3, and @5 are 0.5602, 0.8434, and 0.9398, respectively. As
shown in Figure~\ref{fig:retrieval-comparison}, this selected hybrid
configuration consistently outperforms both the plain semantic and
plain lexical retrieval baselines across all evaluated metrics.

Using this selected configuration, we retrieve and cache the top five
sentences and their scores for each claim--source pair. This cache is
frozen for all subsequent experiments, ensuring the NLI and LLM
verifiers process the exact same inputs so that any difference in
verification performance is independent of retrieval.

\begin{table}[ht]
\centering
\caption{Retrieval performance for different semantic--BM25 weight configurations on the SciFact development set.}
\label{tab:retrieval_weight_selection}
\begin{tabular}{ccccccc}
\hline
\textbf{Semantic-Weight} & \textbf{BM25-Weight} & \textbf{R@1} & \textbf{R@3} & \textbf{R@5} & \textbf{Mean Recall} & \textbf{MRR} \\
\hline
0.0 & 1.0 & 0.4880 & 0.7892 & 0.9277 & 0.7349 & 0.6474 \\
0.1 & 0.9 & 0.5000 & 0.8133 & 0.9398 & 0.7510 & 0.6601 \\
0.2 & 0.8 & 0.5060 & 0.8253 & 0.9398 & 0.7570 & 0.6717 \\
0.3 & 0.7 & 0.5181 & 0.8133 & 0.9458 & 0.7590 & 0.6819 \\
0.4 & 0.6 & 0.5301 & 0.8193 & 0.9458 & 0.7651 & 0.6912 \\
0.5 & 0.5 & 0.5301 & 0.8554 & 0.9518 & 0.7791 & 0.7017 \\
0.6 & 0.4 & 0.5241 & 0.8554 & 0.9518 & 0.7771 & 0.6990 \\
0.7 & 0.3 & 0.5482 & 0.8434 & 0.9458 & 0.7791 & 0.7082 \\
0.8 & 0.2 & 0.5422 & 0.8434 & 0.9458 & 0.7771 & 0.7061 \\
\textbf{0.9} & \textbf{0.1} & \textbf{0.5602} & \textbf{0.8434} & \textbf{0.9398} & \textbf{0.7811} & \textbf{0.7097} \\
1.0 & 0.0 & 0.5542 & 0.8253 & 0.9337 & 0.7711 & 0.7007 \\
\hline
\end{tabular}
\end{table}

\vspace{-5pt}
\subsection{NLI Verification}
\vspace{-5pt}
\label{sec:nli_verification}

For NLI-based verification, we use DeBERTa-v3-base-mnli-fever-anli
\cite{laurer_deberta_v3}, a three-way NLI model built on DeBERTa-v3
\cite{he2021deberta} and fine-tuned on MultiNLI
\cite{williams2018mnli}, FEVER-NLI \cite{nie2019fever}, and ANLI
\cite{nie2020anli}. The model is used zero-shot, without any
fine-tuning on our benchmark, so its performance reflects direct
transfer of pretrained NLI capabilities to claim verification.

For each claim $C$, the top-5 retrieved sentences $E_1, \dots, E_5$
are each paired with the claim to form five NLI inputs $(E_i, C)$. In
every pair, the retrieved sentence $E_i$ is the \textit{premise} and
the claim $C$ is the \textit{hypothesis}, so the model judges whether
the sentence entails the claim, contradicts it, or is neutral toward
it. Each pair is scored independently: the model sees only one sentence
at a time and has no access to the other retrieved sentences.

For each pair, the model outputs a probability for each of the three
classes: \textit{entailment}, \textit{neutral}, and
\textit{contradiction}. We store these probabilities instead of only
the highest-scoring class, as probability also tells us how confident
the model is. For example, a sentence that entails the claim with
probability $0.95$ and one that does so with probability $0.60$ would
both be labeled ``entailment'', even though they provide very
different amounts of support. Each claim therefore ends up with five
sets of NLI probabilities, one per retrieved sentence, each stored with
that sentence's retrieval relevance score $R_i$
(Equation~\ref{eq:relevance}). These are combined into a single
claim-level verdict in the evidence aggregation stage
(Section~\ref{sec:nli_aggregation}).

\vspace{-5pt}
\subsection{NLI Aggregation}
\vspace{-5pt}
\label{sec:nli_aggregation}

The NLI stage produces five probability distributions per claim, one
for each retrieved sentence. We combine them into a single claim-level
verdict using relevance-weighted aggregation. Let
$p_i^{\mathrm{ent}}$, $p_i^{\mathrm{con}}$, and $p_i^{\mathrm{neu}}$
denote the entailment, contradiction, and neutral probabilities for
evidence sentence $E_i$, and let $R_i$ be its retrieval relevance score
(Equation~\ref{eq:relevance}). Each sentence's entailment and
contradiction probabilities are weighted by its relevance and summed
over the five sentences:
\begin{equation}
S_{\mathrm{ent}} = \sum_{i=1}^{5} R_i\, p_i^{\mathrm{ent}},
\qquad
S_{\mathrm{con}} = \sum_{i=1}^{5} R_i\, p_i^{\mathrm{con}}.
\end{equation}
The two scores are then normalized so that they compare support
against contradiction only:
\begin{equation}
P_{\mathrm{ent}} = \frac{S_{\mathrm{ent}}}{S_{\mathrm{ent}} + S_{\mathrm{con}}},
\qquad
P_{\mathrm{con}} = \frac{S_{\mathrm{con}}}{S_{\mathrm{ent}} + S_{\mathrm{con}}}.
\end{equation}
With a decision threshold $\tau$, the verdict is given as follows:

If $P_{\mathrm{ent}} \geq \tau$, the claim is labeled ``Support''; if
$P_{\mathrm{con}} \geq \tau$, it is labeled ``Contradict''; otherwise
it is labeled ``Not Enough Evidence''.\\

Neutral probabilities are deliberately excluded from the
normalization. Consider a claim for which one retrieved sentence
strongly supports it while the other four are unrelated and therefore
mostly neutral. If neutral probability were included, it would dominate
the supporting evidence and wrongly push the claim toward ``Not Enough
Evidence''. Excluding it allows a single directly supporting or
contradicting sentence to determine the verdict without being diluted
by unrelated sentences.

We compare this approach against two simpler alternatives: \textit{Top-1} and \textit{Max-Evidence}. \textit{Top-1} relies only on the most relevant sentence and ignores the other four. In contrast, \textit{Max-Evidence} evaluates all five sentences and independently extracts the maximum entailment probability and the maximum contradiction probability across the entire set, meaning the two strongest signals may originate from different sentences. For both methods, the final verdict is ``Support'' if the entailment probability is at least the threshold $\tau$ and strictly greater than the contradiction probability; ``Contradict'' if the contradiction probability is at least $\tau$ and strictly greater than the entailment probability; and ``Not Enough Evidence'' otherwise.\\

We evaluate the three strategies on the filtered SciFact development
set (166 claims), sweeping $\tau$ from 0.50 to 0.90 in steps of 0.01
for each strategy. Because this set contains only ``Support'' and
``Contradict'' claims, we report binary Macro-F1, averaged over these
two classes, with any ``Not Enough Evidence'' prediction counted as an
error. Relevance-weighted aggregation achieves the highest binary
Macro-F1 (0.7240 at $\tau = 0.52$), compared with 0.6450 for
Max-Evidence (at $\tau = 0.50$) and 0.5710 for Top-1 (at
$\tau = 0.60$). We therefore fix relevance-weighted aggregation with
$\tau = 0.52$ for the final evaluation.

\vspace{-5pt}
\subsection{LLM Verification}
\vspace{-5pt}
\label{sec:llm_verify}

We evaluate three open-source LLMs as verifiers:
Llama 3.1 8B, Qwen3 8B, and Mistral 7B \cite{grattafiori2024llama3,yang2025qwen3,jiang2023mistral7b}.
Like the NLI model, they are used zero-shot, without any fine-tuning on
SciFact, CliniFact, or our benchmark, so their results reflect the
models' pretrained ability to verify claims from retrieved evidence.

For each claim, the LLM receives the claim and the same top-5 evidence
sentences from the frozen retrieval cache, each with its retrieval
relevance score. The prompt instructs the model to use only the
supplied evidence and no outside knowledge, and to answer ``Support''
if the evidence clearly supports the claim, ``Contradict'' if it
clearly contradicts it, and ``Not Enough Evidence'' if it establishes
neither. It tells the model to pay close attention to numerical values, units,
experimental conditions, comparisons, and qualifiers such as
``approximately'' or ``greater than''. The model is also instructed to return only a
JSON object with its verdict. We use temperature zero and the same
prompt and evidence for all three models, so that differences in
performance come from the models themselves.

\vspace{-5pt}
\subsection{Conservative Hybrid Fusion}
\vspace{-5pt}
\label{sec:hybrid}

To combine the NLI and LLM verification outputs, we employ a conservative
agreement-based fusion strategy. The verdicts produced by the NLI and LLM
verifiers are reused directly in the fusion step, without performing an
additional LLM inference. If the NLI and LLM predictions
agree, their shared verdict is returned; otherwise, the claim is classified
as ``Not Enough Evidence''. This strategy prioritizes precision and
avoids accepting a claim when the two independent verification methods
disagree.

\vspace{-5pt}
\section{Experimental Setup}
\vspace{-5pt}
\vspace{-5pt}
\subsection{Models and Implementation}
\vspace{-5pt}

The verification pipeline uses DeBERTa-v3-base-mnli-fever-anli as the NLI
baseline and three open-source large language models: Llama 3.1 8B, Qwen3 8B,
and Mistral 7B. All LLMs are evaluated under the same prompt and evidence
conditions with a temperature of 0, and no model is fine-tuned on the evaluation
data. The LLMs are run locally on an RTX 4080 mobile GPU using Ollama with
quantized model variants. Qwen3 is evaluated with its thinking mode disabled
to maintain a consistent direct-verdict setting across models. Each LLM
response is parsed as JSON to extract the final verdict.

\vspace{-5pt}
\subsection{Evaluation Protocol}
\vspace{-5pt}

The final evaluation is performed on the constructed benchmark of 2,035
claims. The retrieval configuration is frozen at a semantic weight of
0.9 and a BM25 weight of 0.1, with the top five relevant sentences
selected for verification. The NLI aggregation strategy and decision
threshold ($\tau = 0.52$) are also fixed at the configuration selected
on the SciFact development set. The development set is used only for
retrieval and NLI configuration selection and is excluded from the final
evaluation. We evaluate the standalone NLI and LLM verifiers as well as
the corresponding NLI--LLM fusion systems.

\vspace{-5pt}
\subsection{Evaluation Metrics}
\vspace{-5pt}

Macro-F1 is used as the primary classification metric because the
constructed benchmark has an imbalanced class distribution. Macro
Precision, Macro Recall, Accuracy, and per-class Precision, Recall, and
F1 are also reported. Retrieval performance is evaluated on the SciFact
development set using Mean Recall (mean of Recall@1, Recall@3, Recall@5) and Mean Reciprocal Rank (MRR) as mentioned in Section~\ref{sec:hybrid-retrieval}.

\vspace{-5pt}
\section{Results}
\vspace{-5pt}
\label{sec:results}

\vspace{-5pt}
\subsection{Overall Performance}
\vspace{-5pt}

Table~\ref{tab:main} reports accuracy and macro-averaged precision,
recall, and F1 for all seven systems. Because the benchmark is heavily
imbalanced, with 50.4\% ``Not Enough Evidence'', 42.2\% ``Support'',
and 7.5\% ``Contradict'', we use Macro-F1 as the primary metric and not
accuracy. Accuracy counts every claim equally, so always predicting
``Not Enough Evidence'' already gives 50.4\% accuracy but only 0.22
Macro-F1. Macro-F1 weights the three classes equally and combines
precision and recall, so it penalizes systems that do poorly on any
single class.

Among the standalone systems, Qwen3 8B achieves the highest Macro-F1 at
0.7937, with the best macro-precision at 0.8207 and the best
macro-recall at 0.7948. Llama 3.1 8B reaches a Macro-F1 of 0.7204. It
is almost as precise at 0.7987 but has much lower macro-recall at
0.6874. Mistral 7B is weaker on both, with a macro-precision of 0.6523,
a macro-recall of 0.6319, and a Macro-F1 of 0.6388. Llama 3.1 has
slightly higher accuracy than Qwen3, 0.8113 against 0.8079, but a lower
Macro-F1, which shows why accuracy alone is misleading here. The NLI
model reaches only 0.3157 Macro-F1, and its accuracy of 0.3744 is below
the 50.4\% majority-class rate. Its macro-recall of 0.5367 exceeds its
macro-precision of 0.4645 because it labels many claims ``Contradict''
and rarely predicts ``Not Enough Evidence''.

None of the fusion systems improves on its standalone LLM. Fusion
lowers Macro-F1 by 5.4 points for Qwen3, 4.7 for Llama 3.1, and 4.0 for
Mistral, and lowers accuracy by 5.0 to 6.0 points. The drop comes from
macro-recall, which falls by 7.2, 5.1, and 5.0 points, while
macro-precision stays almost unchanged, moving for Qwen3 only from
0.8207 to 0.8234. As Table~\ref{tab:perclass} shows, the precision gain
on ``Support'' and ``Contradict'' is offset by a loss on ``Not Enough
Evidence'', so requiring agreement makes the system miss more claims
without making it more precise overall. Fusion leaves the ranking of
the three LLMs unchanged.

\begin{table}[ht]
\caption{Overall results on the 2,035-claim benchmark. The best value
in each column is in bold.}
\label{tab:main}
\centering
\begin{tabular}{lcccc}
\hline
System & Accuracy & Macro-P & Macro-R & Macro-F1 \\
\hline
NLI (DeBERTa-v3) & 0.3744 & 0.4645 & 0.5367 & 0.3157 \\
Llama 3.1 8B & \textbf{0.8113} & 0.7987 & 0.6874 & 0.7204 \\
Qwen3 8B & 0.8079 & 0.8207 & \textbf{0.7948} & \textbf{0.7937} \\
Mistral 7B & 0.7715 & 0.6523 & 0.6319 & 0.6388 \\
\hline
NLI + Llama 3.1 & 0.7543 & 0.7981 & 0.6369 & 0.6730 \\
NLI + Qwen3 & 0.7479 & \textbf{0.8234} & 0.7233 & 0.7392 \\
NLI + Mistral & 0.7219 & 0.6490 & 0.5824 & 0.5991 \\
\hline
\end{tabular}
\end{table}

\vspace{-5pt}
\subsection{Per-Class Performance}
\vspace{-5pt}
\label{sec:perclass}
Table~\ref{tab:perclass} gives the per-class results. The LLMs perform
similarly on ``Support'' and ``Not Enough Evidence'', with F1 between
0.75 and 0.80 on the first and between 0.82 and 0.85 on the second, but
they differ sharply on ``Contradict'', the rarest class at 7.5\% of the
benchmark. Because Macro-F1 weights all classes equally, this class
largely determines the ranking. Qwen3 reaches a ``Contradict'' F1 of
0.78 at a recall of 0.80, whereas Llama 3.1 reaches 0.51 at a recall of
0.39 and Mistral only 0.31 at a recall of 0.27. Qwen3 is also the most
cautious on ``Support'', with the highest precision at 0.94 but the
lowest recall at 0.63. The NLI model recovers most ``Contradict''
claims, with a recall of 0.89, but at a precision of only 0.15, and it
rarely predicts ``Not Enough Evidence'', the largest class, with a
recall of 0.05.

Fusion affects the classes unevenly. On ``Support'', precision rises,
for example from 0.87 to 0.91 for Llama 3.1, but recall falls by 14 to
20 points, for example from 0.63 to 0.49 for Qwen3. On ``Not Enough
Evidence'', precision falls by 8 to 15 points, for example from 0.76 to
0.68 for Qwen3, because supported claims are moved into this class.
``Contradict'' is barely affected: its F1 stays within 0.01 of the
standalone LLM, as higher precision offsets lower recall. The Macro-F1
loss from fusion comes mostly from ``Support''.

\begin{table}[ht]
\caption{Per-class precision (P), recall (R), and F1 for each system.}
\label{tab:perclass}
\centering
\setlength{\tabcolsep}{3pt}
\resizebox{\linewidth}{!}{%
\begin{tabular}{l ccc ccc ccc}
\hline
 & \multicolumn{3}{c}{Support} & \multicolumn{3}{c}{Contradict}
 & \multicolumn{3}{c}{Not Enough Evidence} \\
System & P & R & F1 & P & R & F1 & P & R & F1 \\
\hline
NLI (DeBERTa-v3) & 0.56 & 0.68 & 0.61 & 0.15 & 0.89 & 0.25 & 0.69 & 0.05 & 0.09 \\
Llama 3.1 8B & 0.87 & 0.74 & 0.80 & 0.75 & 0.39 & 0.51 & 0.78 & 0.93 & 0.85 \\
Qwen3 8B & 0.94 & 0.63 & 0.75 & 0.77 & 0.80 & 0.78 & 0.76 & 0.96 & 0.85 \\
Mistral 7B & 0.75 & 0.82 & 0.79 & 0.37 & 0.27 & 0.31 & 0.84 & 0.80 & 0.82 \\
\hline
NLI + Llama 3.1 & 0.91 & 0.58 & 0.71 & 0.79 & 0.38 & 0.51 & 0.69 & 0.96 & 0.80 \\
NLI + Qwen3 & 0.94 & 0.49 & 0.64 & 0.85 & 0.71 & 0.77 & 0.68 & 0.97 & 0.80 \\
NLI + Mistral & 0.82 & 0.62 & 0.71 & 0.44 & 0.25 & 0.32 & 0.69 & 0.88 & 0.77 \\
\hline
\end{tabular}}
\end{table}

\vspace{-5pt}
\section{Error Analysis}
\vspace{-5pt}
\label{sec:errors}

\vspace{-5pt}
\subsection{NLI Failure Patterns}
\vspace{-5pt}

The NLI model's main weakness is ``Not Enough Evidence'': it labels 978
of the 1,025 ``Not Enough Evidence'' claims as 447 ``Support'' and 531 ``Contradict'', with a recall of only 0.05. These claims all come from CliniFact, and the
failure is specific to that dataset. On SciFact the NLI model reaches an
accuracy of 0.743, close to the LLMs at 0.70 to 0.78, but on CliniFact
it reaches only 0.270, against 0.79 to 0.84 for the LLMs. It labels 697
of the 1,587 CliniFact claims ``Contradict'', although none has that
label.

Two properties of our setup explain this. Each retrieved sentence is
scored independently, and our aggregation compares only entailment
against contradiction (Section~\ref{sec:nli_aggregation}), so a high
neutral probability alone cannot produce ``Not Enough Evidence''. For
61.5\% of the 978 misclassified claims, over 90\% of the weighted NLI
probability is neutral, yet the small remainder decides the verdict.

Two examples show how close to the threshold these decisions are. For
\texttt{clinifact\_5720} (correct label ``Not Enough Evidence''),
neutral accounts for 97\% of the weighted probability and the normalized entailment
score is 0.551, just above $\tau = 0.52$. The NLI model
therefore answers ``Support'', while all three LLMs correctly answer
``Not Enough Evidence''. For \texttt{clinifact\_6762} (correct label
``Support''), neutral accounts for 96\% and the normalized entailment
score is 0.517, just below $\tau$, so the NLI model answers ``Not Enough
Evidence'', while all three LLMs correctly answer ``Support''. In both
cases the verdict is decided by a ratio of very small probabilities
within about 0.03 of the threshold, so a tiny difference produces
opposite errors.

\vspace{-5pt}
\subsection{LLM Verification Errors}
\vspace{-5pt}

The LLMs perform far better but still make errors, and the main error
type differs by model. Qwen3 and Llama 3.1 are cautious: they answer
``Not Enough Evidence'' for claims that do have a ``Support'' or
``Contradict'' label. Qwen3 does so for 289 of the 858 ``Support''
claims (33.7\%), and Llama 3.1 for 206 of them (24.0\%) and for 61 of
the 152 ``Contradict'' claims (40.1\%). Mistral is more error-prone in
the other direction. It labels 57 ``Contradict'' claims ``Support'',
against 32 for Llama 3.1 and only 2 for Qwen3, and it answers ``Support''
for 179 ``Not Enough Evidence'' claims, against 65 for Llama 3.1 and 35
for Qwen3. Missed ``Contradict'' claims are the main reason Llama 3.1 and
Mistral fall behind Qwen3, since they find only 39\% and 27\% of them,
against 80\% for Qwen3.

\vspace{-5pt}
\subsection{NLI--LLM Disagreement and Conservative Fusion}
\vspace{-5pt}

Although both verifiers see the same top-5 sentences, they agree on only
30.8\% of claims for Qwen3, 33.0\% for Llama 3.1, and 38.3\% for
Mistral. When they disagree, the LLM is correct in 76\% to 78\% of cases
and the NLI model in 12\% to 14\%. In 75\% to 89\% of disagreements, the
LLM answers ``Not Enough Evidence'' while the NLI model gives ``Support''
or ``Contradict''. When they agree, the shared verdict is correct for
91.2\%, 88.5\%, and 78.1\% of claims, respectively, above either
verifier's overall accuracy.

The NLI model is nevertheless sometimes the only correct verifier. In
\texttt{scifact\_\allowbreak40\_\allowbreak13497630} (correct label
``Contradict''), the NLI model and Qwen3 answer ``Contradict'', but
Llama 3.1 and Mistral answer ``Not Enough Evidence''. Overall, the NLI
model is correct on 190, 168, and 153 claims that Qwen3, Llama 3.1, and
Mistral get wrong. Picking the correct verifier for every claim would
give Macro-F1 scores of 0.834, 0.824, and 0.783, but this is only an
upper bound, since it requires knowing the correct label in advance.

Conservative fusion does not perform as intended because every
disagreement is mapped to ``Not Enough Evidence''. It fixes only 11, 29,
and 77 incorrect LLM verdicts for Qwen3, Llama 3.1, and Mistral,
respectively, but overrides 133, 145, and 178 correct ones. Consequently, fusion predicts ``Not Enough Evidence'' for 71.8\%, 69.8\%, and 63.7\% of claims, compared with 63.9\%, 60.0\%, and 48.2\% for the standalone LLMs and 50.4\% in the
gold labels. The damage is particularly severe on SciFact since every
SciFact claim in the benchmark has a ``Support'' or ``Contradict'' label,
any disagreement that is converted to ``Not Enough Evidence'' is
necessarily an error. In SciFact accuracy falls by 17--19 points, compared with only 1--3 points on CliniFact. Thus, the conservative agreement rule reduces performance
rather than providing a consistent benefit in this setup.

\vspace{-5pt}
\section{Limitations and Future Work}
\vspace{-5pt}
\label{sec:limitations}

The benchmark is dominated by CliniFact, which contributes 1,587
claims compared with 448 SciFact claims. In addition,
``Contradict'' claims come only from SciFact, while ``Not Enough Evidence''
claims come only from CliniFact. Thus, some of the observed performance
differences maybe due to the differences between the two datasets.
Future work can evaluate the approach on benchmarks with a more balanced
distribution of both datasets and labels. Our experiments also use
zero-shot NLI and compact LLMs with fixed inference settings.
Testing fine-tuned models, additional LLMs, and different model sizes
would provide a broader comparison. Finally, the fusion method uses a
simple agreement rule and treats both verifiers equally. More
sophisticated fusion methods could account for the confidence of each
prediction and the reliability of each verifier.

\vspace{-5pt}
\section{Conclusion}
\vspace{-5pt}
\label{sec:conclusion}

We investigated whether combining NLI with LLM-based verification using a conservative fusion rule can improve claim verification. To provide a strong and consistent evidence set to our verifiers we used a hybrid retrieval approach, that
outperformed both semantic-only and BM25-only retrieval in terms of
MRR and mean Recall. We also evaluated three NLI evidence aggregation
strategies and found that relevance-weighted aggregation performed
best on the development set. On the final benchmark, Qwen3 8B achieved
the strongest standalone performance (Macro-F1 0.7937), substantially
outperforming the NLI baseline (0.3157). Although NLI and
LLMs made different errors, our conservative agreement-based fusion did
not effectively combine their strengths and reduced Macro-F1 for all
three LLMs. These results show that stronger retrieval and diverse
verifiers alone are not sufficient for effective fusion, future
approaches should account for verifier confidence and reliability when
resolving disagreements.\\

\noindent\textbf{Acknowledgements.} We thank Prof. G. L. Priya\inst{1}\orcidID{0000-0001-7759-858X} for guidance and helpful comments on the manuscript.\\
(Source code and datasets:)
\href{https://github.com/plan28-06/Scientific_Claim_Verification_Pipeline}{GitHub Repository},
\href{https://www.kaggle.com/datasets/thedevastator/unlock-insight-into-scientific-claims-with-scifa}{SciFact},
\href{https://github.com/ds4dh/CliniFact/tree/main}{CliniFact}.

\bibliographystyle{splncs04}
\bibliography{references}

@inproceedings{wadden2020,
  author    = {Wadden, David and Lin, Shanchuan and Lo, Kyle and Wang, Lucy Lu and van Zuylen, Madeleine and Cohan, Arman and Hajishirzi, Hannaneh},
  title     = {Fact or Fiction: Verifying Scientific Claims},
  booktitle = {Proceedings of the 2020 Conference on Empirical Methods in Natural Language Processing (EMNLP)},
  pages     = {7534--7550},
  year      = {2020},
  publisher = {Association for Computational Linguistics}
}

@article{wadden2022,
  author  = {Wadden, David and Lo, Kyle and Kuehl, Bailey and Cohan, Arman and Beltagy, Iz and Wang, Lucy Lu and Hajishirzi, Hannaneh},
  title   = {SciFact-Open: Towards Open-Domain Scientific Claim Verification},
  journal = {arXiv preprint arXiv:2210.13777},
  year    = {2022}
}

@article{zhang2025clinifact,
  author  = {Zhang, Boya and Bornet, Alban and Yazdani, Anthony and Khlebnikov, Philipp and Milutinovic, Marija and Rouhizadeh, Hossein and Amini, Poorya and Teodoro, Douglas},
  title   = {A dataset for evaluating clinical research claims in large language models},
  journal = {Scientific Data},
  volume  = {12},
  number  = {1},
  pages   = {86},
  year    = {2025}
}

@inproceedings{reimers2019sentence,
  author    = {Reimers, Nils and Gurevych, Iryna},
  title     = {Sentence-BERT: Sentence Embeddings using Siamese BERT-Networks},
  booktitle = {Proceedings of the 2019 Conference on Empirical Methods in Natural Language Processing and the 9th International Joint Conference on Natural Language Processing (EMNLP-IJCNLP)},
  pages     = {3982--3992},
  year      = {2019}
}

@article{robertson2009probabilistic,
  author  = {Robertson, Stephen and Zaragoza, Hugo},
  title   = {The Probabilistic Relevance Framework: BM25 and Beyond},
  journal = {Foundations and Trends in Information Retrieval},
  volume  = {3},
  number  = {4},
  pages   = {333--389},
  year    = {2009}
}

@inproceedings{wang2020minilm,
  author    = {Wang, Wenhui and Wei, Furu and Dong, Li and Bao, Hangbo and Yang, Nan and Zhou, Ming},
  title     = {MiniLM: Deep Self-Attention Distillation for Task-Agnostic Compression of Pre-Trained Transformers},
  booktitle = {Advances in Neural Information Processing Systems},
  volume    = {33},
  pages     = {5776--5788},
  year      = {2020}
}

@article{he2021deberta,
  author  = {He, Pengcheng and Gao, Jianfeng and Chen, Weizhu},
  title   = {{DeBERTaV3}: Improving {DeBERTa} using {ELECTRA}-Style Pre-Training with Gradient-Disentangled Embedding Sharing},
  journal = {arXiv preprint arXiv:2111.09543},
  year    = {2021}
}

@inproceedings{williams2018mnli,
  author    = {Williams, Adina and Nangia, Nikita and Bowman, Samuel},
  title     = {A Broad-Coverage Challenge Corpus for Sentence Understanding through Inference},
  booktitle = {Proceedings of the 2018 Conference of the North American Chapter of the Association for Computational Linguistics: Human Language Technologies},
  pages     = {1112--1122},
  year      = {2018}
}

@inproceedings{nie2019fever,
  author    = {Nie, Yixin and Chen, Haonan and Bansal, Mohit},
  title     = {Combining Fact Extraction and Verification with Neural Semantic Matching Networks},
  booktitle = {Proceedings of the AAAI Conference on Artificial Intelligence},
  volume    = {33},
  number    = {01},
  pages     = {6859--6866},
  year      = {2019}
}

@inproceedings{nie2020anli,
  author    = {Nie, Yixin and Williams, Adina and Dinan, Emily and Bansal, Mohit and Weston, Jason and Kiela, Douwe},
  title     = {Adversarial {NLI}: A New Benchmark for Natural Language Understanding},
  booktitle = {Proceedings of the 58th Annual Meeting of the Association for Computational Linguistics},
  pages     = {4885--4901},
  year      = {2020}
}

@misc{laurer_deberta_v3,
  author       = {Laurer, Moritz},
  title        = {{DeBERTa-v3-base-mnli-fever-anli}},
  howpublished = {Hugging Face model card, \url{https://huggingface.co/MoritzLaurer/DeBERTa-v3-base-mnli-fever-anli}},
  year         = {2022}
}

@article{grattafiori2024llama3,
  author  = {Grattafiori, Aaron and others},
  title   = {The Llama 3 Herd of Models},
  journal = {arXiv preprint arXiv:2407.21783},
  year    = {2024}
}

@article{yang2025qwen3,
  author  = {Yang, An and others},
  title   = {Qwen3 Technical Report},
  journal = {arXiv preprint arXiv:2505.09388},
  year    = {2025}
}

@article{jiang2023mistral7b,
  author  = {Jiang, Albert Q. and others},
  title   = {Mistral 7B},
  journal = {arXiv preprint arXiv:2310.06825},
  year    = {2023}
}

@inproceedings{kosprdic2024,
  author    = {Ko{\v{s}}prdi{\'c}, Marko and Ljaji{\'c}, Aleksa and Medvecki, Danilo and Ba{\v{s}}aragin, Boris and Milo{\v{s}}evi{\'c}, Nemanja},
  title     = {Scientific Claim Verification with Fine-Tuned NLI Models},
  booktitle = {Proceedings of the 16th International Joint Conference on Knowledge Discovery, Knowledge Engineering and Knowledge Management (IC3K 2024), Volume 3: KMIS},
  pages     = {15--25},
  year      = {2024}
}

@inproceedings{alvarez2024,
  author    = {Alvarez, C. and Bennett, M. and Wang, L.},
  title     = {Zero-shot Scientific Claim Verification Using LLMs and Citation Text},
  booktitle = {Proceedings of the Fourth Workshop on Scholarly Document Processing (SDP 2024)},
  pages     = {269--276},
  year      = {2024}
}

@article{wang2025,
  author  = {Wang, S. and Foulds, J. R. and Gani, M. O. and Pan, S.},
  title   = {LLM-based Corroborating and Refuting Evidence Retrieval for Scientific Claim Verification},
  journal = {arXiv preprint arXiv:2503.07937},
  year    = {2025}
}

@inproceedings{dmonte2026,
  author    = {Dmonte, A. and Oruche, R. R. and Zampieri, M. and Calyam, P. and Augenstein, I.},
  title     = {Claim Verification in the Age of Large Language Models: A Survey},
  booktitle = {Proceedings of the 64th Annual Meeting of the Association for Computational Linguistics (Volume 4: Student Research Workshop)},
  pages     = {15--29},
  year      = {2026}
}

@inproceedings{karpukhin2020dpr,
  author    = {Karpukhin, Vladimir and O{\u{g}}uz, Barlas and Min, Sewon and Lewis, Patrick and Wu, Ledell and Edunov, Sergey and Chen, Danqi and Yih, Wen-tau},
  title     = {Dense Passage Retrieval for Open-Domain Question Answering},
  booktitle = {Proceedings of the 2020 Conference on Empirical Methods in Natural Language Processing (EMNLP)},
  pages     = {6769--6781},
  year      = {2020}
}

@article{bruch2023fusion,
  author  = {Bruch, Sebastian and Gai, Siyu and Ingber, Alon},
  title   = {An Analysis of Fusion Functions for Hybrid Retrieval},
  journal = {ACM Transactions on Information Systems},
  volume  = {42},
  number  = {1},
  pages   = {1--35},
  year    = {2023}
}

@inproceedings{yin2021docnli,
  author    = {Yin, Wenpeng and Radev, Dragomir and Xiong, Caiming},
  title     = {DocNLI: A Large-scale Dataset for Document-level Natural Language Inference},
  booktitle = {Findings of the Association for Computational Linguistics: ACL-IJCNLP 2021},
  pages     = {4913--4922},
  year      = {2021}
}

@article{ji2023survey,
  author  = {Ji, Ziwei and Lee, Nayeon and Frieske, Rita and Yu, Tiezheng and Su, Dan and Xu, Yan and Ishii, Etsuko and Bang, Yejin and Madotto, Andrea and Fung, Pascale},
  title   = {Survey of Hallucination in Natural Language Generation},
  journal = {ACM Computing Surveys},
  volume  = {55},
  number  = {12},
  pages   = {1--38},
  year    = {2023}
}

\end{document}